\documentclass[11pt]{article}

\usepackage{epsfig}

\usepackage{amsmath}
\usepackage{amsfonts}
\usepackage{amssymb}

\title{The Signed Distance Function:\\ A New Tool for 
       Binary Classification}

\author{
Erik M.~Boczko\\
Department of Biomedical Informatics\\ 
Vanderbilt University\\
Nashville, TN 37232\\ 
615-936-6668\\ 
erik.m.boczko@vanderbilt.edu
\and
Todd Young\\ 
Department of Mathematics\\ 
Ohio University \\
Athens, OH 45701\\ 
740-593-1285\\ 
young@math.ohiou.edu}

\newtheorem{fact}{Fact}

\newcommand{\D}[1]{{\mathbb#1}}

\newcommand{\RR}{{\D{R}}}

\begin{document}

\maketitle

\begin{abstract}
From a geometric perspective  
most nonlinear binary classification algorithms, including state of the art 
versions of Support Vector Machine (SVM) and Radial Basis Function 
Network (RBFN) classifiers, and are based on the idea of 
reconstructing indicator functions.
We propose instead to use reconstruction of the signed distance 
function (SDF) as a basis for binary classification.
We discuss properties of the signed distance function that
can be exploited in classification algorithms. 
We develop simple versions of such classifiers and
test them on several linear and nonlinear  problems.
On linear tests accuracy of the new algorithm 
exceeds that of standard SVM methods, with an average of 
50\% fewer misclassifications. 
Performance of the new methods also matches or exceeds that 
of standard  methods on several nonlinear problems including classification
of benchmark diagnostic micro-array data sets. 
\end{abstract}

\noindent
---------------------------------------------------\\
{\small Machine Learning, Microarray Data}\\
---------------------------------------------------

\section{Introduction}
Binary classification is a basic problem in machine learning
with applications in  many fields. Not only does binary 
classification have many potential direct applications, it
is also the basis for many multi-category classification methods.
Of particular interest are the applications in biology and 
medicine. The availability of micro-array and proteomic data 
sets that contain thousands or even tens of thousands
of measurements have particularly made it important to develop good 
classification algorithms, since reliable use of these data could
presumably revolutionize diagnostic medicine. 
Several binary classification algorithms have been developed and 
studied intensely over the past few years, most notable among these 
are the support vector machine (SVM) methods using radial basis 
functions and other functions as kernels. SVM methods have been shown to 
perform reasonably well in classifying micro-array data, demonstrating
that the extraction of useful information from these large data sets
is feasible.

We will begin in the next section with a geometric, 
rather than statistical, statement of
the binary classification problem and our discussion will
be restricted to the context of this geometric viewpoint.
Nonlinear SVM methods, despite their geometric, maximal margin origin,
have been developed based on the idea of reconstructing
``indicator functions", as discussed by Poggio and Smale \cite{PS}. 
RBFN methods are also currently employed in this way. The indicator 
function is an object that encodes only the most primitive geometric 
information. We propose that a potentially better tool 
for classification is the ``signed distance function" (SDF), an 
intrinsically geometric object that has been employed in several 
areas of applied mathematics. The 
geometric properties of the SDF make it 
advantageous for use in classification and we
give examples of how these properties can be exploited.

In this paper we also present preliminary test results for rudimentary 
implementations based on the idea of reconstructing the SDF from 
training data. We demonstrate that these non-optimized SDF-based 
algorithms outperform standard SVM (LIBSVM) methods on average by 50\%
(half as many misclassifications) on linearly separable problems. 
We also present a comparison of SDF classification with SVM method
on the challenging geometric 4 by 4 checkboard problem in which
the SDF-based method performs better.
Finally, on 2 benchmark cancer-diagnosis micro-array data sets  
a nonlinear SDF algorithm
performs just as well or better than highly-developed SVM 
methods. While these results are obviously not conclusive,
they do demonstrate that the SDF paradigm is promising
and worth further investigation.

\section{A Geometric Formulation and a Geometric Tool}

Suppose a set $X \subset \RR^n$ is partitioned by 
$A \subset X$ and its complement $A^c$. The set $A$ might
contain the set of test values that are associated with the 
presence of a disease, while $A^c$ contains those values that are 
not. For applications we may suppose that $A$ is a reasonably
nice set, e.g. has a smooth boundary. The problem of binary
classification is then to determine the set $A$ given a finite sample 
of data, i.e.~for a set of points $\{x_i\}_{i=1}^m$ we know 
whether $x_i \in A$ or $x_i \in A^c$, for each $i$.
The purpose of solving this problem is obviously predictive 
power; given a point $x \in X$ 
that is not among the given data $\{x_i\}_{i=1}^m$, we wish to determine 
if $x \in A$. In this paper we only consider this geometric 
formulation of the problem. While this formulation is obviously
somewhat restrictive, it allows for geometric analysis and leads
to a new class of methods that may be useful in some applications.

One way to approach this problem mathematically that is common
among nonlinear classifiers is to consider 
the {\em indicator function} of $A$, $\iota_A:X \rightarrow \{-1,1\}$, defined by
\begin{equation}
  \iota_A(x) = 
       \begin{cases}
       \: 1 & \text{if $x \in A$,}\\
       -1 & \text{if $x \in A^c$.}
       \end{cases}
\end{equation}
The known information is represented by $\{(x_i,\iota_A(x_i)\}_{i=1}^m$ 
and problem of binary classification is equivalent to reconstructing
$\iota_A(x)$ from the data.

Current methods of binary classification such as the 
SVM methods and RBFN methods
work by attempting to approximate the indicator function
$\iota_A$ by regression over the known data, $\{(x_i,\iota_A(x_i)\}$. 
This was not the conceptual origin of the SVM, i.e. finding a 
separating plane of maximal margin, but is the basis
for nonlinear ``kernel trick" algorithms and efficient linear 
implementation as pointed out in \cite{PS}.
In practice, the SVM constructed functions are smooth and $\pm 1$ are
not the only values in the range, thus $x$ is interpreted as in $A$ if the
constructed function is positive at $x$ and as being in $A^c$ if 
it is negative.

Rather than using the indicator function $\iota_A$, we propose
using the {\em signed distance function} (SDF) of $A$, denoted $b_A(x)$
which is the distance to the boundary, $\partial A$, of $A$ if $x \in A$ 
or minus the 
distance to $\partial A$ if $x \in A^c$, i.e. 
\begin{equation}
  b_A(x) = 
       \begin{cases}
       \; d(x,\partial A) & \text{if $x \in  A$,}\\
         - d(x,\partial A) & \text{if $x \in A^c$,}
       \end{cases}
\end{equation}
where $d$ is a metric (distance function).

Knowledge of $b_A$ is obviously 
sufficient to fully determine the set $A$, it carries more information
than $\iota_A$, and it has some smoothness. These 
and other properties can be used in classification and 
thus the SDF has advantages over the indicator function as a 
basis for binary classification algorithms. 
In fact we argue
that SDF based classification, because it is more geometrical, 
is conceptually a more faithful generalization
of the original SVM concept than existing nonlinear 
(kernel trick) SVM implementations.

Binary classification based on the SDF can work in 
much the same way as indicator function based
classification; one  attempts to approximate the function $b_A$
using only the given data. If the value of $b_A$ is positive
at a test point $x$, then $x$ is 
predicted to be in $A$ and if the value is negative, $x$
is predicted to be in $A^c$. The approximation of
$b_A$ can proceed similarly to that of $\iota_A$, i.e.~by
various forms of regression (including SVM and RBFN regression). 
A practical difference is that $\iota_A$ 
is given explicitly at the data points, whereas $b_A$ at the 
data points must be derived from the data. We investigate
simple methods for doing this and show that they give 
reliable results. We also show that properties
of $b_A$ can be used to refine those estimates. The complexity
of this task is no worse than that needed to perform the regression
itself, hence no computational performance is sacrificed.


 
\section{Signed Distance Functions}

In some places $b_A$ is called the  oriented boundary distance 
function or the oriented distance function.
Proofs of the following facts about $b_A$ can be found in \cite{DZ}.
\begin{fact}
The function $b_A$ is Lipschitz continuous, with Lipschitz constant $1$.
\end{fact}
In other words, $|b_A(x) - b_A(y)| < |x - y|$, holds for all $x,y \in X$.
This implies that $b_A(x)$ is differentiable almost everywhere, i.e,
$Db_A(x)$ exists except on a set of zero measure. It also 
implies that $b_A(x)$  belongs to the Sobolev space 
$W^{1,p}_{loc}$ for any $1 \le p \le \infty$.

\begin{fact}\label{grad}
If $b_A$ is differentiable at a point $x$, then there exists a unique $P x \in \partial A$
such that $b_A(x) = |x - Px|$ and 
$$
     \nabla b_A(x) =  \frac{Px -x}{b_A(x)}.
$$ 
\end{fact}
In the case it is unique, $Px \in \partial A$ is called the 
projection of $x$ onto $\partial A$. In particular, 
$|\nabla b_A(x)| = 1$ and $D b_A(x)$ points from $x$ toward $Px$.

\begin{fact}
Let $A$ be a subset of $\RR^n$ with nonempty boundary $\partial A$. Then
$b_A$ is a convex function if and only if $\overline{A}$ is a convex set.
If $A$ is convex, then $b_A$ is differentiable everywhere in $A^c$.
\end{fact}

\begin{fact}\label{locdiff}
If $\partial A$ is of smoothness class $C^k$, i.e. it is $k$ times continuously 
differentiable, $k\ge 1$, 
then for each $y \in \partial A$ 
there is a neighborhood $V(y)$ of $y$ on which $b_A$ is a $C^k$ function.
\end{fact}
If $\partial A$ is $C^1$, then at any point $y \in \partial A$
we can define the unit normal vector $n(y)$.
\begin{fact}\label{normal}
Suppose $\partial A$ is of smoothness class $C^1$ and $n(y)$ is
Lipschitz continuous. At any point $y \in \partial A$, let 
$V(y)$ be as in Fact~\ref{locdiff}. Then for any $x \in V(y)$, 
$x = Px + b_A(x)n(Px)$. 
\end{fact}
In particular, $Px -x$ and $D b_A(x)$ are normal to $\partial A$ at $Px$.

\begin{fact}
Let $H_y$ denote the mean curvature of $\partial A$ at a point $y \in \partial A$.
Then wherever $H_y$ exists it satisfies
$$
                           H_y = \Delta b_A(x),
$$
where $\Delta$ is the usual Laplace operator. 
\end{fact}
The function $b_A$ has been used in various branches of applied math 
such as free boundary problems \cite{CZ, ES, OF, Seth} and grid generation 
for finite-element methods \cite{PSt}. It is intimately related
to flow by mean curvature \cite {DZ} and occurs in the solution
of certain Hamilton-Jacobi partial differential equations \cite[p.~163]{E}.
The geometric nature of the SDF connects it to well developed 
areas of geometry
and analysis that can be expected to provide tools for both refinement 
and analysis of 
SDF based classification methods.

\section{SDF Classifiers}

\subsection{Preliminary algorithms}

In the SDF paradigm the input training data are marked 
as to class, but they do not come marked with the values $b_A(x_i)$, 
and hence these need to be approximated. A naive SDF algorithm 
then consists of two simple steps, with an optional third refinement 
step.
\begin{itemize} 
\item Approximate $b_A$ at the training data $\{x_i\}_{i=1}^m$. Denote these 
      approximations by $\{b_i\}_{i=1}^m$ 
\item Approximate $b_A$ by a function $B_D(x)$ on the entire domain through 
      regression on $D :=\{(x_i,b_i)\}_{i=1}^m$.
\item Use the constructed function $B_D$ and properties of $b_A$ to improve 
      the estimates $\{b_i\}$ and iterate.
\end{itemize}
We now detail preliminary algorithms for these three steps and 
point to those areas that we consider important for further investigation.

\subsection{Estimating $b_A$ at the data}

Let $d$ denote a metric on $X \subset \RR^n$. Usually we
will let $d$ be the Euclidean metric, but we will
also consider weighted distances.
A reasonable and simple first approximation of $b_A$ at 
$\{x_i\}_{i=1}^m$ is given by
$$
 b_i = P'(x_i) \equiv \iota_A(x_i) \cdot \min_{j \neq i} 
     \{  d(x_i,x_j): \iota_A(x_j) \neq \iota_A(x_i) \},
$$
i.e. the signed projection onto the (finite) data of opposite type.
It is clear that $b_i$ has the correct sign and is a bound on 
$b_A(x_i)$, i.e.
\begin{equation}\label{upperbound}
        |b_A(x_i)| \le |b_i|.
\end{equation}

It is easy to show by counter-example that obtaining more precise, 
yet rigorous, bounds on $b_A(x_i)$ would require some
assumptions on the shape of $A$. However, we can make some 
heuristic improvements in $\{b_i\}$. For instance, consider
$$
       \tilde{b}_i = b_i - b_A(P' x_i).
$$
Then we have
$$
        |b_A(x_i)| \le |\tilde{b}_i| \le |b_i|.
$$
Now suppose that $x_i \in A$ and $y_i \in A^c$
where $y_i$ is the closest data point to $x_i$ in $A^c$ and $x_i$
is the closest data point in $A$ to $y_i$. If $\partial A$ is
situated half way between $x_i$ and $y_i$ and is normal to $y_i-x_i$,
then $b_A(x) =  d(x,y)/2$.
Let $c_i$ denote the approximated signed distance of $y_i$ to the
boundary, then we have:
$$
         b_A(x_i) = \frac{1}{2} b_i = b_i - \frac{1}{2} c_i.
$$
Then for any $x_i$ and $y_i = P' x_i$ we define
\begin{equation}\label{bprime}
        b'_i = b_i - .5 c_i
\end{equation}
where $b_i$ is the first approximation of $b_A(x_i)$ and $c_i$
is the first approximation of $b_A(y_i)$. It can be demonstrated
that, on average, $b'_i$ is a better approximation of $b_A(x_i)$ than $b_i$.

\subsection{Linear classifier}

We will say that a binary classification problem in the context
of our geometric formulation is {\em linearly
separable} if $\partial A$ is a hyperplane in $\RR^n$. 
In this
case $b_A(x)$ will be a linear function whose zero set is
$\partial A$. To use the SDF for a linearly separable problem
one would seek a linear function as $B_D(x)$, i.e.
$$
    y = \ell(x) =  w \cdot x + c = w_1 x_1 + w_2 x_2 + \ldots + w_n x_n + c
$$
that fits the data $\{x_i,b_i)\}_{i=1}^m$. The most obvious choice for
this approximation is to use the linear least squares approximation
if $m > n$ or projection (pseudoinverse) if $m < n$,
for which there are highly developed algorithms. In the linear tests
below, we have used linear least squares regression.

From \S 2, we  have $| \nabla b_A(x) | = 1 $ wherever it exists so 
we could let
$$
     |D\ell(x)| = |w| = 1
$$
be a constraint in the linear least squares
approximation. 

\subsection{Nonlinear classifier}

If the problem is not linearly separable, then $b_A(x)$ will
be a nonlinear function. To approximate it we should use some
form of nonlinear regression on $\{(x_i,b_i)\}_{i=1}^m$.
There are many options for nonlinear regression that could
be used, including SVM and RBFN regression algorithms. 
One simple, yet appealing, choice for the nonlinear regression
is the least squares regression discussed in \cite{PS}.
We implemented this approach in the nonlinear tests
reported below and so we recall it.

Let $K: X \times X \rightarrow \RR$ be a kernel 
that is symmetric ($K(x,y) = K(y,x)$) and positive-definite, i.e.,
$$
     \sum_{i,j=1}^k c_i c_j K(x_i,x_j) \ge 0,
$$
for any $k$, any $x_1, \ldots, x_k$ and any $c_1, \ldots, c_k$.
In the tests below we use the Gaussian
$$
     K(x,y) = e^{-d(x,y)^2/2 \sigma^2},
$$
which is symmetric and positive definite. Let $\mathbf{K}$ be the square, 
positive-definite matrix with elements $K_{i,j} = K(x_i,x_j)$.
Let $\mathbf{I}$ be the identity matrix and $\mathbf{b}$ be
the vector with coordinates $b_i$. Let $\gamma >0$ and let $\mathbf{c}$ 
be the solution of the linear system of equations:
\begin{equation}\label{lsr}
      ( \mathbf K + m \gamma \mathbf{I}) \mathbf{c} = \mathbf{b}.
\end{equation}
This problem is well-posed since $( \mathbf K + m \gamma \mathbf{I})$
is strictly positive-definite and the condition number will
be good provided that $m \gamma$ is not too small. The number $\gamma$
can be viewed as a smoothing parameter. Given the solution vector 
$\mathbf{c}$, the approximation, $B_D(x)$, of $b_A(x)$ is given by
$$
    B_D(x) = \sum_{i=1}^m c_i K(x_i,x).
$$

The choices of $\sigma$ and $\gamma$ in this approximation are
discussed in \cite{PS} and elsewhere. In tests
discussed below, results were insensitive to
$\gamma$ within a fairly large range. Good choices for 
$\sigma$, which we found by cross-validation, were on the 
order of the mean inter-data distance.

We emphasize that other regression methods could be used in
connection with SDF-based and should be investigated.

\subsection{Iteration}

There are some possibilities for refining the initial data
approximation $\{b_i\}$. One idea for nonlinear problems would
is to let 
$$
b'_i = B_D(x_i)
$$
where $B_D$ is the regression obtained from $\{(x_i,b_i)\}$.
Then one could take $b'_i$ as a refinement of $b_i$ and then 
run the regression again.
(For linear regression, this would repeat the original results.)
We note that iterations of this type are related to the matrix 
eigenvalue problem for which there are well developed numerical
techniques and which are amenable to stability analysis.

Another possibility is to use Facts~\ref{grad}~and~\ref{normal} above
to refine $b_i$. From these facts it is reasonable to project
$ y_i - x_i$ onto $D f(x_i)$. This approach is particularly
appealing for problems that are known a priori to be linearly
separable. We have  successfully implemented this iterative approach 
in the linear tests below, using the projection onto $w = D\ell$
to refine $\{b_i\}_{i=1}^m$.

\section{Test Results}

\subsection{Linearly separable problems}

We applied both non-iterative and iterative forms of the 
linear regression signed distance classifier to three types of 
distributions: uniform, normal and skewed. In all of these tests 
the linear SDF classifier decisively outperforms the linear 
classifiers in the LIBSVM package as well as the linear Lagrangian SVM
\cite{MM} and the Proximal SVM \cite{FM}.

A linearly separable problem can be transformed by a linear
change of coordinate to the problem where $\partial A = \{x_n=0\}$.
Thus we use this problem for our tests. We performed the tests
for $n = 2$. Data in
the half space $x_n >0$ we labeled as in $A$ and data with $x_n <0$
we labeled as in $A^c$.

In the uniform distribution
tests we let the domain be the square $[-1,1] \times [-1,1] \subset \RR^2$.
For the normal distribution we used the standard normal distribution at the
origin in $\RR^2$. For the skewed distribution, we randomly choose points
in the square $[0,1]^2 \subset \RR^2$ using the density
$\rho(x_i) = 1/\sqrt{x_i}$, then scaled them affinely to $[-1,1]^2$.

In the tests, we considered training sets of size from $m = 10$ to $m=10,000$. 
For each $m$ in the range we classified 50 distributions
of $m$ points. In each test we used a
test set consisting of 4000 points selected randomly according to
the distribution type being tested.

\begin{figure}[hbtp]
\centerline{\hbox{\psfig{figure=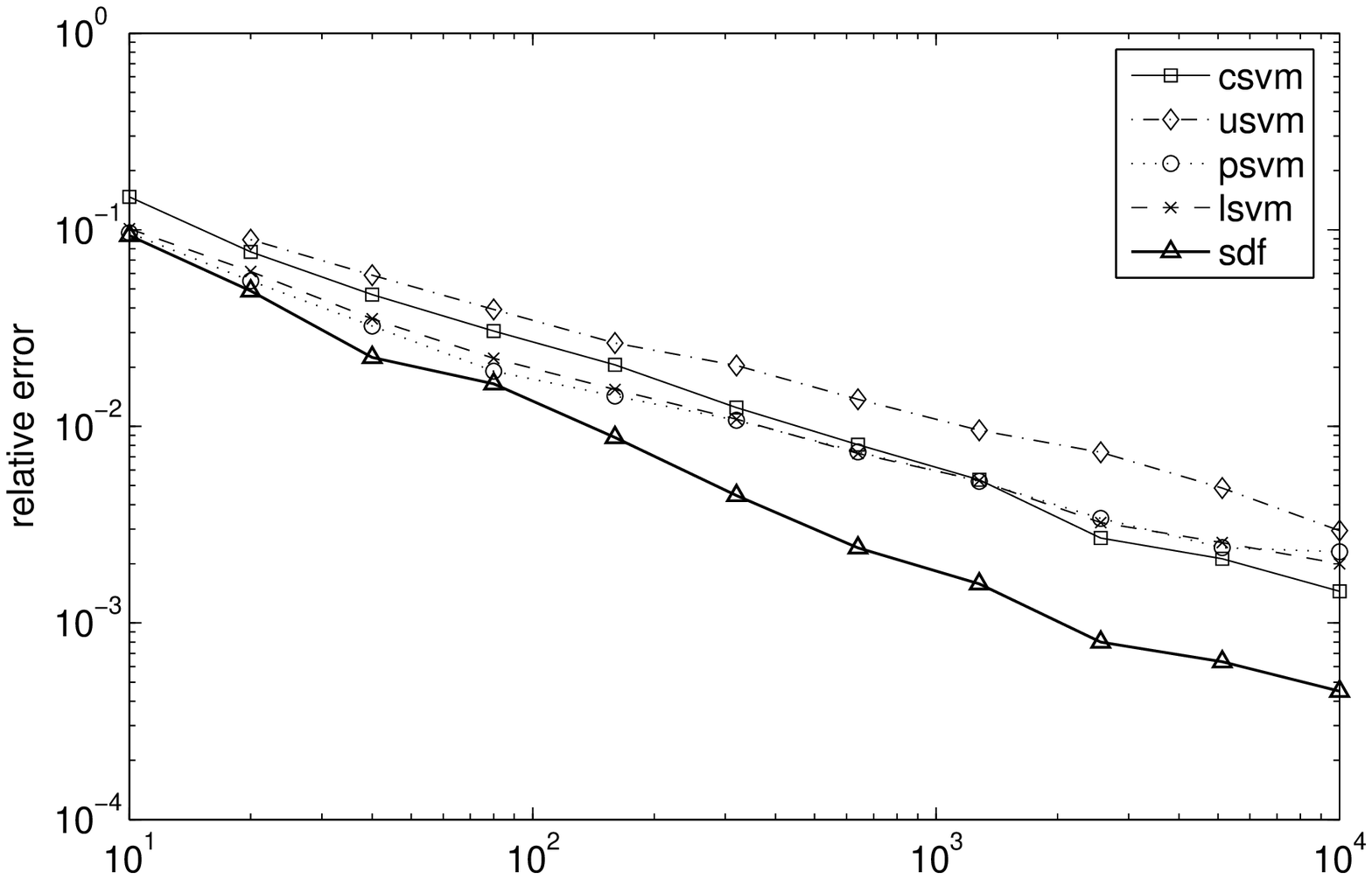,height=2.0in,width=3.5in}}}
\vspace*{-.1cm}
\centerline{\hspace*{-.3cm}\hbox{\psfig{figure=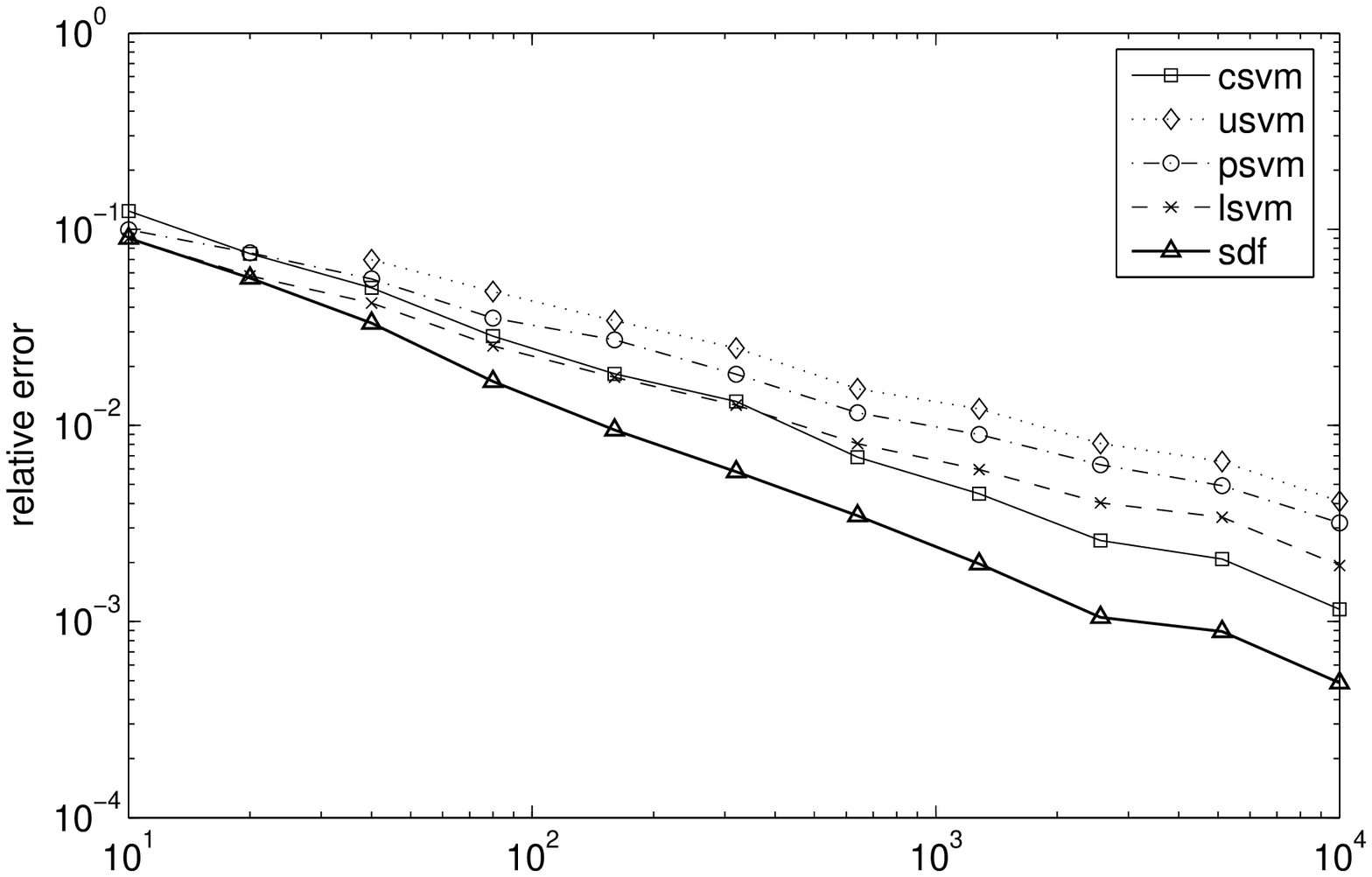,height=1.9in,width=3.1in}}}
\centerline{\hbox{\psfig{figure=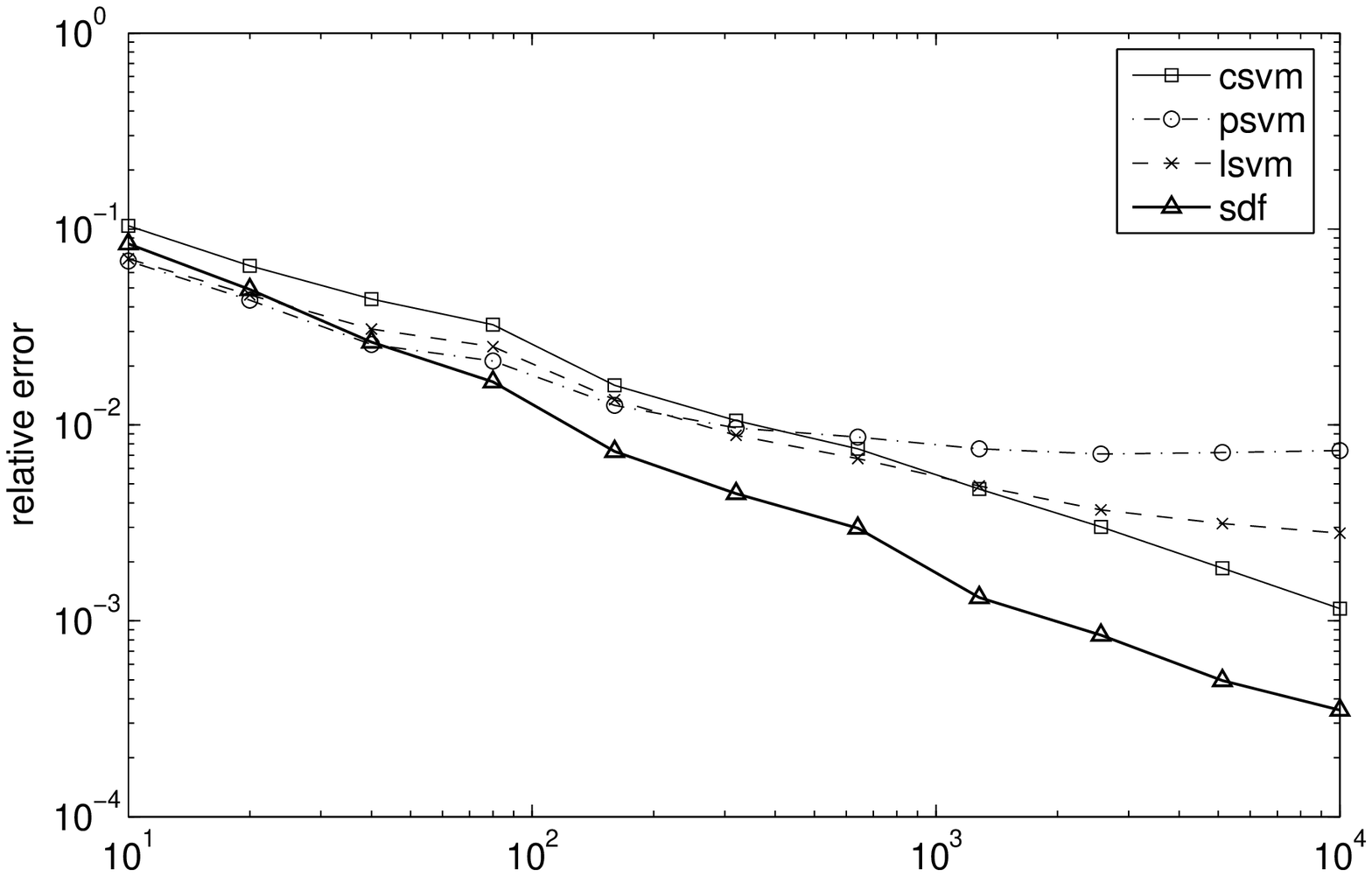,height=2.0in,width=3.5in}}}
\vspace*{-.2cm}            
\caption{A comparison of classifiers on linearly 
separable data. In (a) the data are uniformly distributed, 
in (b) the data are normally distributed, and in (c) the 
data are from a skewed distribution. The $x$-axis is $\log(m)$, 
where $m$ is the number of data (training) points. Each point in the 
graph represents an average over 50 independent tests.}
\label{fig:lin}
\end{figure}

In Figures~\ref{fig:lin} we show comparisons 
of the SDF linear classifiers with the linear classifier from the
LIBSVM along with the Lagrangian SVM \cite{MM} and Proximal SVM \cite{FM}. 
In this plot \verb&csvm& and \verb&usvm& are routines from the LIBSVM
package, \verb&psvm& is the Proximal SVM and \verb&lsvm& is the 
Lagrangian SVM algorithm. 
In these tests  the iterated SDF method was iterated 5 times.
The iterated SDF method gave a 10\% to 15\% decrease in the error
over the non-iterated SDF method. The Lagrangian method was also
iterated 5 times. The LIBSVM package methods automatically iterate.
In our trials, the number of iterations for the LIBSVM methods 
increased with $m$ and varied from 10 to 5000 iterations.

It can be seen that the SDF
classifier has noticably smaller errors than either SVM method over a large
range of number of training points $m$.
Averaged over all 550 tests, the SDF-based classification produced 52\%
fewer misclassifications than the best SVM (LIBSVM c-SVM) method.
(Average $\approx 98\%$ correct vs. $ \approx 96\%$ correct).

\subsection{The $4\times 4$ Checkerboard Problem}

\begin{figure}[hbt]
    \begin{center}
        \includegraphics[width=0.5\linewidth]{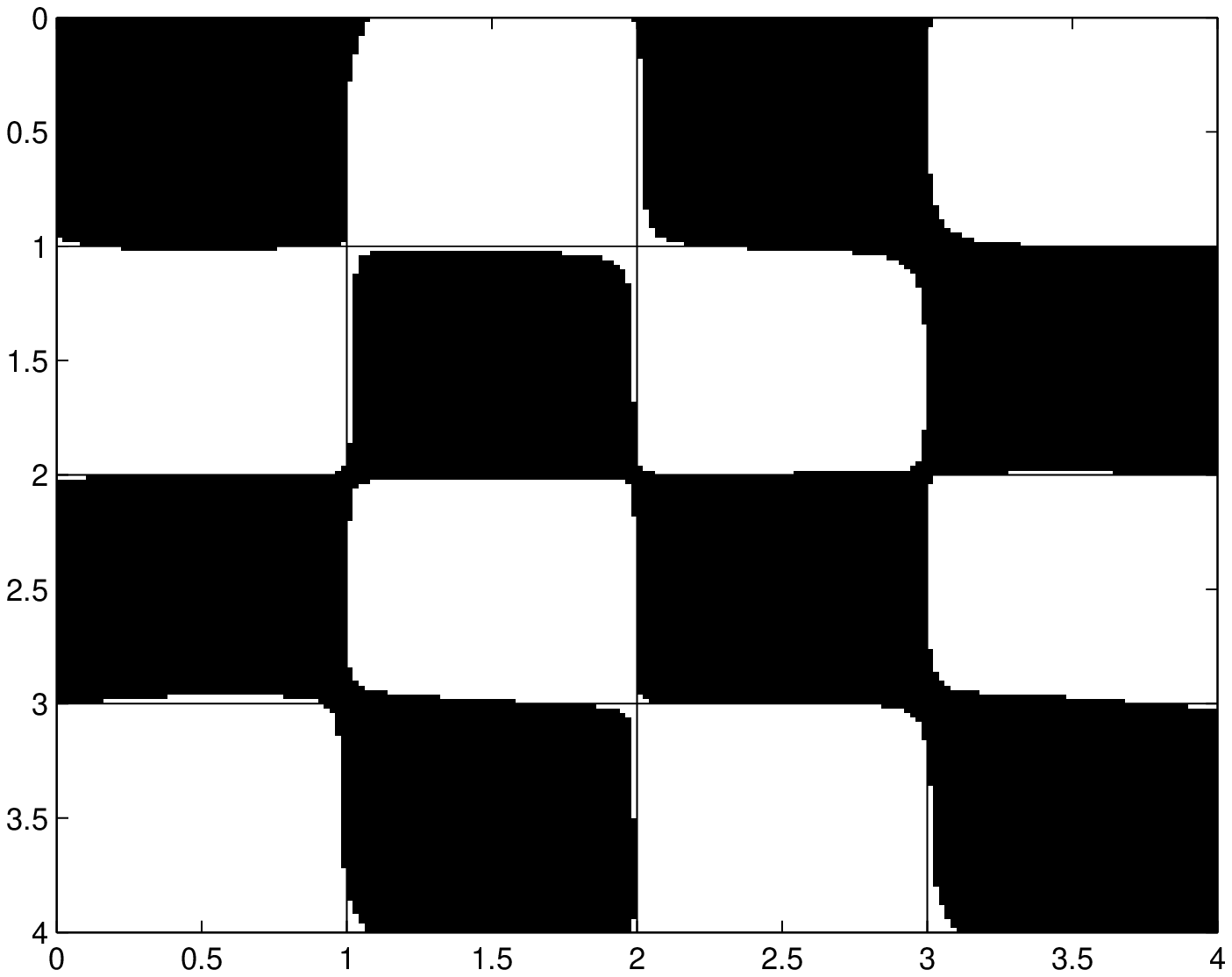}
    \end{center}
    \caption{The 4 by 4 checkerboard problem classified by a nonlinear
             SDF-based classifier.}
    \label{fig2}
\end{figure}

There are several
benchmark nonlinear problems, but perhaps the prototype is the 4
by 4 checkerboard. This geometric problem is interesting because
it is known to be difficult. In this test a square is partitioned
into 16 equal sub squares with alternate squares belong to two
distinct types, black or white (Figure~\ref{fig2}). Following
\cite{MM}, we used 1,000 randomly selected points in each
training set and 40,000 grid points as the test set.

\begin{figure}[hbt]
    \begin{center}
        \includegraphics[width=0.6\linewidth]{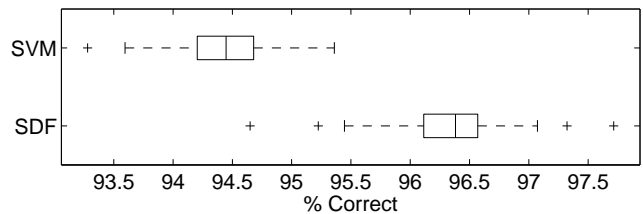}
    \end{center}
\caption{Comparison of performance by the LIBSVM package and a 
         SDF-based method on the four by four checkerboard problem.}
    \label{fig3}
\end{figure}

In Figure~\ref{fig3} we show the results of
applying an SDF-based classification scheme and the standard LIBSVM
package to 100 independent training set.
We used the nonlinear least squares regression with parameters
$\sigma$ and $\gamma$ found by cross-validation on the training
sets. Parameters for the SVM classification were chosen by precisely
the same process. 

Note that the SDF-based method produces better
results than the LIBSVM package. The mean correct \%
and standard deviation for the SDF method were 96.3\%
and .46\%. 
For the SVM method the mean correct \%
was 94.5\% 
with a standard deviation of .36\%.

The standard deviation for our SDF method is slightly higher
than that for the LIBSVM package. This is perhaps an artifact
of our naive implementation of a SDF-based method.

We note that~\cite{MM} reported $97.1\%$ accuracy on this problem
using a Lagrangian SVM with 100,000 iterations.

\subsection{Micro-array data sets}

We compare the nonlinear SDF classifier
with existing studies of SVM performance
on two standard micro-array data sets involving cancer diagnosis.
The first set, called Prostate--tumor, consists of 102 micro-array samples
with  10,510 measurements each. Each patient represented by a 
sample was diagnosed independently for the presence of a prostate
tumor. Of the 102 samples, 52 were from patients with a prostate
tumor \cite{Sin}.
The second data set DLBCL consists of 77 samples with 5470
variables each. Samples were taken from patients 
diagnosed for a lymphoma. Of the 77 patients, 58 had Diffuse large
B-cell lymphomas and 19 had follicular lymphomas \cite{Shi}.

With the two data sets, we tested the SDF nonlinear classifier with 
least squares regression on the data sets using Leave One Out Cross 
Validation (LOOCV). Accuracy of the LOOCV tests
are shown in Table~1, with comparison of reported LOOCV performance
from \cite{stat1} using a variety of SVM methods and the k-nearest
neighbor (KNN) method. The percentages
reported are the percentages of correctly identified samples.
It is important to point out that~\cite{stat1} showed that their 
error estimates did not change depending on the cross validation technique, 
and hence LOOCV is a robust estimate of performance
when applied to these data.

\begin{table}[hbt]
\centerline{
\begin{tabular}{|c|ccc|}
\hline
      Data set &  KNN    &  SVM      &  SDF      \\ \hline
      Prostate--tumor  &  $85\%$ & $92.2\%$  & $94.1\%$  \\ 
        DLBCL          &  $87\%$ & $97.4\%$  & $97.4\%$  \\
\hline
\end{tabular}
}
\caption{Results from applying SDF-based classification to
the benchmark DLBCL and Prostate--tumor micro-array data sets compared 
with the performance of SVM methods and the k-nearest neighbor method
reported in \cite{stat1}.
All results are for Leave-one-out cross validation. 
}
\label{cancer}
\end{table}

As seen in Table~\ref{cancer}, the performance of the SDF classifier
matches that of the SVM method on the DLBCL data and exceeds it
on the Prostate--tumor data set.

In preliminary tests we found that performance on several 
LOOCV subsets once again was unaffected 
by $\gamma$ over a broad range $10^{-4}$--$10^{-12}$ .
Based on this we simply set $\gamma = 10^{-7}$ for the rest 
of the tests. Values for $\sigma$ 
were determined within each loop of the LOOCV process
based on mean interdata distances for the subset. 
For the DLBCL data 
values of $\sigma$ were approximately $2.5 \times 10^4$ and 
$\sigma$ was generally about $3.4 \times 10^3$ 
for subsets of the Prostate--tumor set. For both of these data sets
the optimal results are actually robust with respect to changes
in the values of $\sigma$. 

In these tests we  used a weighted distance
$$
   d_a(x,y) = | a \cdot (x-y) |,
$$
where $| \cdot |$ is the usual Euclidean norm and $a$ is a vector of weights.
Distance functions of this type were shown to be effective in high 
dimensional binary classification problems in \cite{BYD}.
Specifically, we took $a$ to be the absolute values of the correlation coefficients
relating each variable to the indicator on the data set. This was recalculated
in the LOOCV process for each subset, independent of the excluded sample. 

We note that in an independent set of experiments reported in 
\cite{BYWX}, the nonlinear SDF-based classifier was compared to 
KNN, RBFN and SVM classifiers on five other cancer data sets. 
The following microarray data
sets are involved: The \emph{Breast Cancer} data set~\cite{Wes01} consists
    of 49 tumor samples with 7129 human genes each. There are two
    different response variables in the data set: one describes
    the status of the estrogen receptor (ER), and the other one
    describes the status of the lymph nodal (LN), which is an
    indicator of the metastatic spread of the tumor. Of the 49
    samples, 25 are ER+ and 24 are ER-, 25 are LN+ and 24 are LN-.
    The \emph{Colon Cancer} data set~\cite{Alon} consists
    of 40 tumor and 22 normal colon tissues with 2000 genes each.
    The \emph{Leukaemia} data set~\cite{Gol} consists of
    72 samples with 7129 genes each. Each patient represented by
    a sample has either acute lymphoblastic leukemia (ALL) or
    acute myeloid leukemia (AML). Of the 72 samples, 47 are ALL
and 25 are AML.

We tested the four classifiers in 100 independent trials on each
of the data sets. In each trial, the data were divided randomly
into a training set and a test set in a ratio of 2:1.
We used Gaussian kernel functions for RBFN, SVM and SDF classifiers. For
simplicity, we did not use any heuristic for the distance metrics
of KNN, using the Euclidean distance. We claim that the
classifiers are comparable in this setting since they are under
exactly the same condition: (i) They share the same training set
and test set in each trial, (ii) SVM and SDF share the same
$\gamma=10^{-7}$, (iii) SVM uses the weighted kernel matrix
returned by SDF in each trial, (iv) SVM and RBFN use the same
$\sigma$, which is computed in each trial as the root mean square 
distance (RMSD) of the training data.

\begin{table}[hbt]
    \begin{center}
        \begin{tabular}{|c|cccc|}
            \hline
            \multicolumn{1}{|c|}{Data Set} &\multicolumn{1}{c}{KNN} &\multicolumn{1}{c}{RBFN} &\multicolumn{1}{c}{SVM}
            &\multicolumn{1}{c|}{SDF} \\
            \hline
            Breast cancer, ER &.0912 &.0912 &.0869 &.0869 \\
            Breast cancer, LN &.2400 &.2425 &.2106 &.2100 \\
            Colon cancer &.2200 &.2143 &.1700 &.1662 \\
            Leukaemia &.0146 &.0321 &.0167 &.0167 \\
            \hline
        \end{tabular}
    \end{center}
    \caption{Comparison of misclassification ratios averaged over 100
    trials on randomly divided data.} \label{tab3}
\end{table}

Table~\ref{tab3} shows the test error rates averaged over the 100
independent trials for each classifier. KNN with $k = 1, 9, 3, 5$
neighbors achieves the best (in the averaging sense)
generalization performance for the breast cancer data (ER), breast
cancer data (LN), colon cancer data, and the leukemia data, respectively. 
Note that in actual use, $k$ would have to be determined in
some unbiased way from the training data only. We note again
that the naive SDF method matches or beats the SVM method on all
data.

\section{Discussion}

In order to make the SDF paradigm competitive with indicator function
based classification, the main need seems to be for more accurate, 
yet efficient, ways of obtaining an approximation of $b_A(x_i)$. 
In the scheme we used in these tests, we simply search 
the entire data set for the closest point of the opposite type. In 
the worst case this takes $m^2$ operations, which is easily
within the realm of practical computations. 
Increasing the accuracy of the approximation
is a more difficult issue and should involve deeper geometric 
information from the data set. 

In addition to better determination of $\{b_i\}$, use of other
methods of nonlinear regression, including SVM and RBFN regression, 
with SDF-based classification
should be explored.
Another area for future exploration is development of iterative
methods for the nonlinear classifier. We have described two
possible procedures for this iteration and implemented one
in a linear setting.

Smale and coworkers have been developing methods for rigorous
estimates for the least squares regression algorithm outlined in \S2.5.
They produce these estimates in the framework of Reproducing Kernel
Hilbert Spaces which have been shown to be isomorphic to certain
Sobolev spaces \cite{SZ}, including the space $H^1_{loc}$, to which
signed distance function are known to belong \cite{DZ}. 
The estimates could be used in our context
if the accuracy of the initial estimates $\{b_i\}_{i=1}^m$ are known. 
For the naive method of determining these values, an upper bound 
is given by (\ref{upperbound}) and we hope to obtain better bounds 
under assumptions on $A$. Other geometric methods for approximating 
$\{b_i\}_{i=1}^m$ should lend themselves to rigorous analysis depending 
on the methods.
Perhaps for the case of iterative methods, the iteration process could 
be linked to known results in PDE and related functional analysis.
Such a link would make an extremely rich arena of knowledge available 
for the purposes of estimates.

The above estimates of the approximated SDF should
not only result in a overall reliability measure of the method, 
but should
provide for any given test point  an estimate of the 
distance of that test point from the decision surface.
Combining this with statistical knowledge of the underlying 
application could provide a very natural ``level of confidence" measure 
for any given test data. Such estimates would be especially
useful in the context of biomedical applications.

There are concrete mathematical reasons why the SDF is a better basis
than the indicator function for use in classification. The SDF
is fundamentally geometric and this connects it solidly to
geometric and analytical tools and methods. 
In preliminary tests, we have shown that a naive, non-optimized 
implementation of SDF-based classification is non-trivially more
accurate than standard methods on geometric problems. In preliminary tests on 
nonlinear, high dimensional and noisy data, we have 
demonstrated that a non-optimized implementation of 
SDF is at least as accurate as current, standard SVM methods. 
These observations and results indicate that the SDF paradigm has the potential
to be the basis for more accurate binary
classification algorithms in many contexts.

\end{document}